\documentclass{article}
\usepackage{multirow}
\usepackage{microtype}
\usepackage{graphicx}
\usepackage{subfigure}
\usepackage{booktabs} 

\usepackage{hyperref}

\usepackage[accepted]{icml2025}

\usepackage{amsmath}
\usepackage{amssymb}
\usepackage{mathtools}
\usepackage{amsthm}

\usepackage[capitalize,noabbrev]{cleveref}

\theoremstyle{plain}

\theoremstyle{definition}

\theoremstyle{remark}

\usepackage[textsize=tiny]{todonotes}

\icmltitlerunning{Rate, Explain and Cite (REC): Enhanced Explanation and Attribution in Automatic Evaluation by LLMs}

\begin{document}

\twocolumn[
\icmltitle{Rate, Explain and Cite (REC): Enhanced Explanation and Attribution in Automatic Evaluation by Large Language Models}



\icmlsetsymbol{equal}{*}

\begin{icmlauthorlist}
\icmlauthor{Aliyah R. Hsu}{yyy,comp}
\icmlauthor{James Zhu}{comp}
\icmlauthor{Zhichao Wang}{comp}
\icmlauthor{Bin Bi}{comp}
\icmlauthor{Shubham Mehrotra}{comp}
\icmlauthor{Shiva K. Pentyala}{comp}
\icmlauthor{Katherine Tan}{comp}
\icmlauthor{Xiang-Bo Mao}{comp}
\icmlauthor{Roshanak Omrani}{comp}
\icmlauthor{Sougata Chaudhuri}{comp}
\icmlauthor{Regunathan Radhakrishnan}{comp}
\icmlauthor{Sitaram Asur}{comp}
\icmlauthor{Claire Na Cheng}{comp}
\icmlauthor{Bin Yu}{yyy}
\end{icmlauthorlist}

\icmlaffiliation{yyy}{UC Berkeley}
\icmlaffiliation{comp}{Salesforce AI Platform}

\icmlcorrespondingauthor{Aliyah R. Hsu}{aliyahhsu@berkeley.edu}
\icmlcorrespondingauthor{James Zhu}{james.zhu@salesforce.com}

\icmlkeywords{Explainable AI, Trustworthy AI, Large Language Models}

\vskip 0.3in
]



\printAffiliationsAndNotice{Work done during an internship at Salesforce.}  

\begin{abstract}
LLMs have demonstrated impressive proficiency in generating coherent and high-quality text, making them valuable across a range of text-generation tasks. However, rigorous evaluation of this generated content is crucial, as ensuring its quality remains a significant challenge due to persistent issues such as factual inaccuracies and hallucination. This paper introduces three fine-tuned general-purpose LLM auto-evaluators, REC-8B, REC-12B and REC-70B, specifically designed to evaluate generated text across several dimensions: faithfulness, instruction following, coherence, and completeness. These models not only provide ratings for these metrics but also offer detailed explanation and verifiable citation, thereby enhancing trust in the content.
Moreover, the models support various citation modes, accommodating different requirements for latency and granularity. Extensive evaluations on diverse benchmarks demonstrate that our general-purpose LLM auto-evaluator, REC-70B, outperforms state-of-the-art LLMs, excelling in content evaluation by delivering better quality explanation and citation with minimal bias. Our REC dataset and models are available at \url{https://github.com/adelaidehsu/REC}.
\end{abstract}

\section{Introduction}
\label{sec:intro}
\begin{figure*}[t]
    \centering
    \includegraphics[width=\textwidth, height=\textheight, keepaspectratio]{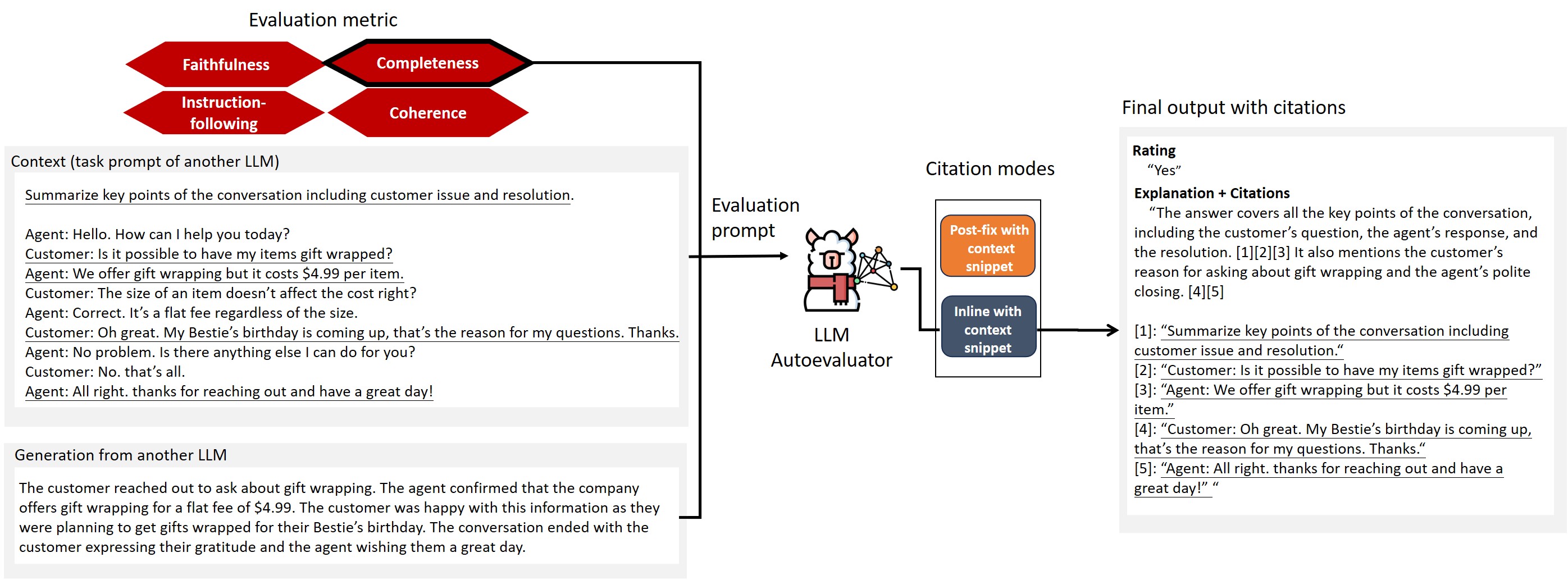}
    \caption{Illustration of the REC framework for content quality evaluation. The auto-evaluator takes in context~(a task prompt from another LLM), generation from another LLM, and a user-specified evaluation metric~(i.e., completeness). The auto-evaluator outputs rating, explanation with citation according to a user-specified citation mode~(i.e., inline with context snippet). Citation are extracted verbatim from context as underlined. For details of the alternative citation mode and the evaluation prompt, see Appendix~\ref{sec:appendix_content_quality_citation}.}
    \label{fig:abcd_illustration_rec}
\end{figure*}

Large Language Models~(LLMs) have demonstrated impressive capabilities in generating high quality coherent text and are deployed in applications for various text-generation tasks~\citep{Brown2020GPT3, Chowdhery2022PaLMSL, Achiam2023GPT4TR, Touvron2023Llama2O}. In order for LLMs to provide up-to-date information or to perform knowledge-intensive tasks~\citep{Lewis2020RetrievalAugmentedGknowledgeintense}, Retrieval Augmented Generation~(RAG) system~\citep{chen-etal-2017-wikirag, Borgeaud2021ImprovingLMRAG, Izacard2022FewshotLWRAG, Guu2020pretrainRAG} has been widely used. RAG involves first retrieving documents that are relevant to the generation task from an external knowledge source, and performing the generation using a knowledge-augmented prompt. However, it is of paramount importance to ensure that the generated text is reliable and can be trusted by a human, as LLMs often suffer from factual inaccuracies and hallucination of knowledge~\citep{Ji2022SurveyOHhalu, Zhang2023SAC3RHhaludetect, Shuster2021RetrievalARhaluconvo}. Towards this goal, we propose that the generated text needs to be evaluated along various dimensions shown below:

\begin{itemize}
    \item \textbf{Faithfulness:}Is the LLM generated response factually correct given the context?
    \item \textbf{Instruction Following:} Does the LLM generated response follow the instructions provided in the prompt?
    \item \textbf{Coherence:} Is the LLM generated response coherent?
    \item \textbf{Completeness:} Does the LLM generated response include all the necessary details?
    \item \textbf{Citation:} If a LLM generated response is factually correct, can we provide evidence for where that response came from?
\end{itemize}

In this paper, we introduce fine-tuned models for the evaluation tasks listed above that can form the basis for all trust-related evaluations for generative AI applications. The models were fine-tuned to not only provide ratings for the metrics but also explanations + citation for why it rated the generation so. 
An example of how our models operate for content quality evaluation can be found in Figure~\ref{fig:abcd_illustration_rec}.

Our models are the \emph{first} to enable citation in both automatic evaluation~(i.e., content quality citation) and general task output~(i.e., general RAG citation) with scalability and efficiency, while prior work often focus on providing solely rating and explanation in automatic evaluation~\citep{zhu2023judgelm, li2023generative-auto-j, wang2023shepherd, Vu2024Foundational-auto-rater}, or require iterative prompting~\citep{Sun2023TowardsVT, cao-wang-2024-verifiable, Ye2023EffectiveLL-AGREE} or human annotation~\citep{malaviya2024expertqa} for general citation generation.
Our contributions include the following:
\begin{itemize}
    \item A novel general-purpose LLM auto-evaluator that comes in three sizes: 8B, 12B and 70B. Our LLM auto-evaluators can generate improved quality \textbf{R}ating, \textbf{E}xplanation and \textbf{C}itation~(\textbf{REC}), with little to no trade-off in general instruction task performance.~\footnote{All evaluations are limited to English datasets.}
    
    \item A curated dataset for citation and explanation fine-tuning to facilitate future research, which is the \emph{first} public dataset containing both \emph{content quality citation}~\footnote{Citation to support LLM generated explanation for content quality evaluation. See Section~\ref{sec:rec_model_capability} for more details.} and \emph{RAG citation}.
    
    \item A single model that can perform different modes of citation to cover the trade-off between latency and granularity of citation. 
\end{itemize}

We provide extensive evaluations of our REC models on diverse benchmarks evaluating LLMs' (1) \emph{RAG citation} capability: ALCE~\citep{Gao2023EnablingLL-alce}, and ExpertQA~\citep{malaviya2024expertqa}; (2) \emph{content quality citation} capability: human evaluation on ABCD summarization~\citep{chen-etal-2021-action}; (3) general capabilities: RewardBench~\citep{RewardBench}, and LLM-AggreFact~\citep{tang-etal-2024-minicheck-llm-aggrefact}; and (4) cognitive bias: CoBBLEr~\citep{koo-etal-2024-benchmarking}, against 7 state-of-the-art~(SOTA) LLMs, such as GPTs and Claude. We show that our REC models outperform state-of-the-art baselines on these benchmarks, and that the baselines often struggle to provide accurate citation, undermining their reliability. 

\section{REC Model Citation Capabilities}
\label{sec:rec_model_capability}

In this section, we discuss the two main citation tasks, content quality citation and RAG citation, that the REC models support, with the understanding that the models can also be used for general purposes, such as question answering~(QA) and summarization.

We illustrate the content quality citation task in Figure~\ref{fig:abcd_illustration_rec}, where the auto-evaluator performs content quality evaluation on another LLM's generation according to a specified metric, and provides rating and explanation with citation for verifiability as an output.
When not in an evaluation setting, our models can perform general RAG citation, where we cite from retrieved articles and tag the citation with an answer generated by another LLM, as illustrated in Figure~\ref{fig:rag_illustration}.

Our models support various citation modes dependent on the citation tasks: post-fix citation~(RAG only), inline citation~(RAG only), post-fix citation with context snippet~(for both content quality evaluation and RAG), and inline citation with context snippet~(for both content quality evaluation and RAG). Output examples of different citation modes can be found in Appendices~\ref{sec:appendix_content_quality_citation} and~\ref{sec:appendix_rag_citation}.

These different modes were designed to accommodate different trade-offs between latency and granularity of citation. For instance, the post-fix citation mode would be the fastest as there is no need for our models to generate claims or snippets. It simply has to generate the reference to the cited context. On the other hand, the inline citation with context snippet mode has to place the citation inline with the generated response and also point to a snippet within the corresponding cited context. This mode is the most granular, but can increase latency as it needs to generate more output tokens.

\section{Related Work}
\label{sec:related_work}
\paragraph{Verifiable Text Generation}
Providing a reference to the attributable source of information~\citep{Liu2023EvaluatingVI, Gao2023EnablingLL-alce} has been proposed as one of the approaches to mitigate hallucination and improve the factuality of LLM generation. Such work fall into the general \emph{RAG citation} setting defined in our work, where the aim is to generate verifiable content through citation of provided articles. Initial efforts fine-tune LLMs using human-written examples~\citep{Nakano2021WebGPTBQ} or machine-generated examples verified by humans~\citep{Menick2022TeachingLM}, but their privately maintained training data limits further research. With the advent of more capable LLMs~\citep{Achiam2023GPT4TR, Jiang2023Mistral7}, most existing work rely on zero-shot prompting or few-shot prompting to cite articles during generation~\citep{Kamalloo2023HAGRIDAH, Gao2023EnablingLL-alce, Liu2023EvaluatingVI}, although the quality
of their generated citation leaves much room for
improvement~\citep{malaviya2024expertqa}. Other work utilize an additional natural language inference~(NLI) model to add citation~\citep{Gao2022RARRRA, Chen2023PURREE}. To improve upon prior work, we propose a fine-tuned LLM that can generate better-grounded responses supported with citation of various modes, with the fine-tuning dataset released to the public to facilitate future research and no additional NLI models needed. 

\paragraph{General-purpose auto-evaluators}
Collecting human annotations to evaluate LLM is not only costly and time consuming, but hard to replicate~\citep{Ouyang2022TrainingLM, Zheng2023JudgingLW, chiang-lee-2023-large}, and as a result, LLMs become a natural automated proxy to evaluate LLM capabilities on various benchmarks~\citep{bai-benchmark, Bubeck2023SparksOA, vicuna2023, fu2023gptscore, liu-etal-2023-geval, wang-etal-2023-chatgpt}. Existing LLM auto-evaluators often judge LLM outputs by expressing ``preference'' over outputs from a reference model~\citep{dubois2024lengthcontrolled, alpaca_eval, yuchen2024wildbench}, or by providing rating and explanation according to a user-defined metric as a direct assessment~\citep{li2023generative-auto-j, wang2023shepherd, pandalm2024}. Closer to our work, \citet{jiang2024tigerscore} and \citet{xu-etal-2023-instructscore} fine-tune LLMs to generate rating, explanation, and a detailed analysis to pinpoint errors in a response evaluated. Unlike prior work, our fine-tuned model is a more generalizable auto-evaluator since it provides \emph{content quality citation} for both good and bad evaluated responses, in addition to rating and explanation. This not only allows users to be able to diagnose where errors are from but provides supporting evidence when a generation is good, which are both important to establish trust in a model.

\section{Data Collection}
\label{sec:data_collection}
To fine-tune our general-purpose LLM auto-evaluator, we meticulously collect a mixture of data encompassing a broad spectrum of LLM capabilities~(Section~\ref{sec:data-mixture}), and we denote our collected dataset as REC-Data. Due to a lack of publicly available content quality citation datasets, we leverage synthetic data generation using an LLM to curate such data~(Section~\ref{sec:synthetic-data-gen}) to facilitate the research community. We perform rule-based automatic quality check followed by a unified task formatting to post-process our data for instruction fine-tuning~(Section~\ref{sec:postprocess}).
\subsection{Task Types}
\label{sec:data-mixture}
To enhance the explainability in the automatic evaluation of our LLM auto-evaluator, while maintaining its general instruction-following capability, we gather datasets from a diverse range of task types~(See detailed REC-Data distribution in Appendix~\ref{appendix:rec-data}). These tasks represent essential capabilities that an advanced auto-evaluator LLM should possess:
\begin{itemize}
\item Pairwise Evaluation: Compare two responses at the same time and express a preference according to evaluation criteria. 
\item Pointwise Evaluation: Evaluate specific aspects of a response independently according to evaluation criteria and provide a rating.
\item Open-ended Evaluation: Evaluate a response independently and provide a free-form explanation, usually to support either pairwise or pointwise evaluation.
\item Citation: Evaluate a response alongside the context, and provide citation for verifiable answer attribution.
\item General Instruction: Generate a response as instructed~(no evaluation tasks in this type), such as summarization and QA.
\end{itemize}

\subsection{Synthetic Data Generation}
\label{sec:synthetic-data-gen}
We leverage synthetic data generation for some of the tasks, including pointwise evaluation and citation. To ensure the quality of synthetic data, we leverage a powerful instruction-tuned model based on Llama-3.1-70B to generate both pointwise evaluation data and citation data for RAG and content quality.~\footnote{Agreement between this model's judgment and human labelers' judgment is 86\%, on par with the agreement between human labelers.} For pointwise evaluation, the model rates every response on each of the 4 metrics listed in Section~\ref{sec:intro}. Example prompts used to generate the synthetic data are provided in Appendix~\ref{appendix:synthetic_prompt}.

\subsection{Post-processing and Unified Task Format}
\label{sec:postprocess}
Knowing that LLM-generated outputs are susceptible to hallucination, we further post-process the synthetic data to ensure the quality. Specifically, we filter synthetic LLM outputs that are either in wrong JSON output formats, or with invalid citation by checking whether the citation is extracted verbatim from the context. Our final REC-Data has in total around 140k datapoints, each containing a task prompt and a completion.

\section{Model Training}
\label{sec:model}
Our general-purpose LLM auto-evaluator is available in three sizes, REC-8B, REC-12B, and REC-70B.
REC-12B is instruction fine-tuned from Mistral-Nemo\footnote{\url{https://mistral.ai/news/mistral-nemo}}, and REC-8B and REC-70B are instruction fine-tuned from Llama-3.1-8B and Llama-3.1-70B \cite{dubey2024llama3herdmodels}, respectively. We trained REC-8B and REC-12B to assess the generalizability of the REC approach to different model architectures in similar model sizes. We adopted supervised fine-tuning~(SFT) to optimize the models.
To accommodate GPU memory constraints: we filtered examples where the combined length of a prompt and a response exceeded 6,144 tokens. We used Low-Rank Adaptation~(LoRA) during fine-tuning~\cite{hu2021loralowrankadaptationlarge}, with a rank of $r=256$ for REC-8B and REC-12B, and $r=64$ for REC-70B. We applied 4-bit quantization for REC-70B during initialization. All models were trained using a learning rate of $1\times10^{-4}$, with data shuffled and for a single epoch. REC-8B and REC-12B were trained with a batch size of 2 and a gradient accumulation factor of 8, whereas for REC-70B we used a batch size of 1 with the same gradient accumulation factor. All models were trained on eight NVIDIA H100 GPUs, each with 80GB of memory. The total training time was approximately 4.5 hours for REC-8B and REC-12B, and 5.5 hours for REC-70B.

\section{Evaluation}
\label{sec:evaluation}
In this section, we evaluate our REC models on diverse benchmarks assessing LLMs' (1) \emph{RAG citation} capability: ALCE~\citep{Gao2023EnablingLL-alce}, and ExpertQA~\citep{malaviya2024expertqa}; (2) \emph{content quality citation} capability: human evaluation on ABCD summarization~\citep{chen-etal-2021-action}; (3) general capabilities: RewardBench~\citep{RewardBench}, and LLM-AggreFact~\citep{tang-etal-2024-minicheck-llm-aggrefact}; and (4) cognitive bias: CoBBLEr~\citep{koo-etal-2024-benchmarking}, each of which is an important measure to a general-purpose LLM auto-evaluator. We compare our results against 7 SOTA LLMs, including Misral-7B (\texttt{Mistral-7B-Instruct-v0.2})~\citep{jiang2023mistral7b}, Mistral-Nemo (\texttt{Mistral-Nemo-Instruct-2407}), Llama-3.1-70B~\citep{grattafiori2024llama3herdmodels}, Claude-3-Opus (\texttt{claude-3-opus-20240229}), GPT-3.5 (\texttt{gpt-3.5-turbo}), GPT-4 (\texttt{gpt-4-turbo})~\citep{openai2024gpt4technicalreport}, and GPT-4o (\texttt{gpt-4o}).

\begin{table}[t]
\small
  \centering
  \begin{tabular}{lcc}
    \hline
    & \textbf{AutoAIS} & \textbf{FActscore} \\ \hline
    \textbf{Mistral-7B} & 26.39 & 84.03 \\
    \textbf{Mistral-Nemo} & 41.39 & 85.78 \\
    \textbf{Llama-3.1-70B} & 52.11 & 86.42 \\
    \textbf{Claude-3-Opus} & 59.84 & 87.16 \\
    \textbf{GPT-3.5} & 57.65 & 86.27 \\
    \textbf{GPT-4} & 60.18 & 87.31 \\
    \textbf{GPT-4o} & 60.86 & 87.03 \\
    \textbf{REC-8B} & 51.36 & 85.81 \\
    \textbf{REC-12B} & \underline{62.90} & \underline{88.22} \\
    \textbf{REC-70B} & \textbf{66.52} & \textbf{89.05} \\ \hline
  \end{tabular}
  \caption{General RAG citation quality and answer correctness results on ExpertQA. The best and the second best performance for each metric are in bold and underlined separately.}
  \label{tab:expertqa}
\end{table}

\begin{table}[t]
\small
  \centering
  \begin{tabular}{lccc}
    \hline
    & \textbf{Fluency} & \textbf{Correct} & \textbf{Citation F1}\\ \hline
    \textbf{Mistral-7B} & 64.49 & 15.12 & 42.94 \\
    \textbf{Mistral-Nemo} & 61.35 & 18.03 & 41.32 \\
    \textbf{Llama-3.1-70B} & \underline{72.01} & 19.09 & 40.18 \\
    \textbf{Claude-3-Opus} & 67.53 & 12.11 & 35.30 \\
    \textbf{GPT-3.5} & 51.20 & 19.02 & \textbf{50.87} \\
    \textbf{GPT-4} & 57.98 & 20.08 & \underline{47.09} \\
    \textbf{GPT-4o} & 56.29 & \textbf{22.47} & 40.50 \\
    \textbf{REC-8B} & 68.20 & 18.01 & 38.77 \\
    \textbf{REC-12B} & 64.94 & 19.60 & 43.63 \\
    \textbf{REC-70B} & \textbf{73.08} & \underline{20.52} & 41.73 \\ \hline
  \end{tabular}
  \caption{Average fluency, answer correctness and citation quality across datasets on ALCE. The best and the second best performance for each metric are in bold and underlined separately.}
  \label{tab:alce}
\end{table}

\subsection{RAG citation \& Correctness}
\label{subsec:rag-citation-eval}
The two benchmarks discussed in this section evaluate LLMs' ability to generate RAG citation in the \emph{inline} mode.
\paragraph{ExpertQA}
ExpertQA~\cite{malaviya2024expertqa} is designed to evaluate LLMs' ability to generate accurate and well-attributed responses in technical and high-stakes fields, such as medicine and law. It was developed by involving experts from 32 different fields, who contributed 2,177 domain-specific questions based on their knowledge. In total, 484 participants helped curate these questions. LLMs were then used to generate responses to the questions, followed by human experts evaluating the quality of these responses on several criteria, including factual correctness, completeness of attribution, source reliability, and informativeness.
We adopt the metrics used in its paper: AutoAIS~\cite{Gao2022RARRRA}, which is similar to citation recall, and FActscore~\cite{min-etal-2023-factscore} which measures the percentage of generated claims that are factual. 

As shown by the zero-shot results in Table~\ref{tab:expertqa}, our REC-12B and REC-70B achieve the highest AutoAIS and FActscore scores, and our REC-8B is also able to outperform other similar-sized models~(Mistral-7B and Mistral-Nemo),  
which not only indicates their better citation quality but also the generalizability of their attribution capability across domains.

\paragraph{ALCE}
The ALCE~(Automatic LLMs' Citation Evaluation) benchmark~\citep{Gao2023EnablingLL-alce} is designed to assess the ability of LLMs to generate text with accurate and relevant citation, focusing on three key dimensions: fluency, answer correctness, and citation quality.
The ALCE benchmark is built on three datasets: ASQA~(a short-answer question dataset), QAMPARI~(which provides lists of correct answers), and ELI5~(a long-form question-answer dataset). Note that fluency is not measured in QAMPARI because of the nature of its answers being in lists. For each dataset, ALCE evaluates how well a model generates text that is not only fluent and accurate but also properly supported by citation. The citation quality evaluation includes citation precision~(ensuring all cited sources are relevant) and citation recall~(ensuring all necessary sources are cited).~\footnote{For a detailed definition of the fluency and answer correctness metrics, please refer to \citet{Gao2023EnablingLL-alce}.}

From Table~\ref{tab:alce}, we see that our REC-70B model ranks highly in fluency and answer correctness. Although the GPT models excel in citation quality in these datasets, the much smaller REC-12B and REC-70B models outperform most of the competitors. When taking the three metrics together and considering the overall average performance, REC-70B achieves an average score of 45.11, the highest among the compared models.
Since all criteria are important in generating a good response with accurate citations~\footnote{For why the three metrics are all important to a verifiable LLM generation, see Appendix~\ref{sec:appendix_alce_remark}.}, this highlights the robustness and balance of REC-70B across fluency, correctness, and citation quality. REC offers a well-rounded solution for different types of questions, making it highly reliable for real-world applications where both citation quality and answer correctness are essential. 

\begin{table*}[t]
\small
\centering
\resizebox{\textwidth}{!}{%
\begin{tabular}{l|ccc|ccc|ccc}
\hline
\multirow{2}{*}{} & \multicolumn{3}{c|}{\textbf{Instruction-following}} & \multicolumn{3}{c|}{\textbf{Completeness}} & \multicolumn{3}{c}{\textbf{Faithfulness}} \\ \cline{2-10} 
                       & \textbf{Rate Acc} & \textbf{Explain Acc} & \textbf{Citation (F1)} & \textbf{Rate Acc} & \textbf{Explain Acc} & \textbf{Citation (F1)} & \textbf{Rate Acc} & \textbf{Explain Acc} & \textbf{Citation (F1)} \\ \hline
\textbf{Mistral-7B}             & \textbf{1.00}        & \textbf{1.00}     & 0.05          & 0.92   & 0.97   & 0.04   & 0.97  & 0.96  & 0.08   \\
\textbf{Mistral-Nemo}           & 0.96        & 0.96     & 0.51          & 0.93   & 0.95   & 0.11   & 0.95  & 0.95  & 0.29   \\
\textbf{Llama-3.1-70B} & 0.98        & 0.98     & \textbf{0.73} & 0.96   & 0.97   & 0.54   & 0.97  & 0.96  & 0.37   \\
\textbf{Claude3-Opus}           & \textbf{1.00}        & \textbf{1.00}     & 0.67          & \textbf{0.98}   & \textbf{0.99}   & 0.46   & \textbf{0.98}  & 0.97  & 0.59   \\
\textbf{GPT-3.5}                & 0.88        & 0.99     & 0.48          & 0.88   & 0.58   & 0.21   & 0.78  & \textbf{0.99}  & 0.29   \\
\textbf{GPT-4}                  & 0.97        & 0.97     & 0.38          & 0.96   & 0.96   & 0.38   & 0.96  & 0.96  & 0.28   \\
\textbf{GPT-4o}                 & 0.95        & 0.95     & 0.35          & 0.93   & 0.93   & 0.56   & 0.96  & 0.71  & 0.59   \\
\textbf{REC-8B}               & \textbf{1.00}        & 0.98     & 0.48          & 0.94   & 0.95   & 0.60   & 0.95  & 0.98  & 0.59   \\
\textbf{REC-12B}               & \textbf{1.00}        & \textbf{1.00}     & 0.50          & 0.94   & 0.95   & 0.62   & 0.94  & 0.98  & 0.60   \\
\textbf{REC-70B}               & \textbf{1.00}        & \textbf{1.00}     & 0.50          & 0.96   & 0.96   & \textbf{0.67}   & 0.97  & 0.98  & \textbf{0.63}   \\ \hline
\end{tabular}
}
\caption{Human evaluation results on ABCD summarization content quality evaluation task.} 
\label{tab:ABCD_human_label_metrics}
\end{table*}

\subsection{Rating \& Explanation \& Citation}
\label{subsec:content-quality-citation-eval}
The benchmark discussed in this section evaluates LLMs' ability to generate content quality citation in the \emph{inline with context snippet} mode.
\paragraph{ABCD}
To evaluate our LLM auto-evaluator's citation capability more comprehensively, besides examining its general \emph{RAG citation} capability on QA tasks as reported in Section~\ref{subsec:rag-citation-eval}, we evaluate the \emph{content quality citation} capability on a summarization task, the ABCD dataset~\cite{chen-etal-2021-action}, in this section.

The ABCD dataset~\cite{chen-etal-2021-action} contains customer support conversation transcripts.
We generate summaries by prompting an open-source LLM, Mistral-7B~\citep{Jiang2023Mistral7}, with: ``Summarize knowledge from transcripts after they've ended, including the customer issue and resolution.'' to form (dialogue, summary) pairs for evaluation. LLM auto-evaluators then take as input the summarization task prompt, and a (dialogue, summary) pair, to generate rating, explanation, and citation according to the metrics defined in Section~\ref{sec:intro}.
Note that we use a held-out set of 50 examples in this experiment so there exists no data overlap with the ABCD subsets used during training. 

Unlike the general \emph{RAG citation} benchmarks evaluated in Section~\ref{subsec:rag-citation-eval}, which contain human annotated citation groundtruth, there exists no well-established benchmarks to evaluate \emph{content quality citation}. Hence, we seek human annotations to serve as a reference standard in this task.
The machine outputs were labeled by 12 machine learning experts, each of whom has a graduate degree in computer science or related fields.
They were asked to evaluate the correctness of each of the rating, explanation, and citation generated by the machine.
For rating and explanation evaluation, the annotators were asked to provide ``Yes/No'' labels to determine the correctness of the response.
For citation evaluation, they were asked to to provide manually-written citation for the claims in the generated explanation to serve as the groudtruth citation.~(Details of the human labeling guideline can be found in Appendix~\ref{sec:appendix_human_label}).
Human evaluation on the Coherence metric was not performed because it is conceptually challenging to define what should be cited in a coherent response (e.g., citing an entire paragraph adds little interpretive value). Citation for coherence are more meaningful in negative cases where parts of the response are incoherent. To ensure a balanced evaluation with both positive and negative cases across metrics, we decided to exclude Coherence from the ABCD summarization content quality task.

We assign two labelers for each model and find the average inter-rater agreement is around 95\%.
We report the average accuracy for rating and explanation across the two sets of labels, and we compute F1 score between machine-generated citation and the intersection of the two sets of manual citation.
From Table~\ref{tab:ABCD_human_label_metrics}, we see that most LLM auto-evaluators obtain high rating and explanation accuracies, indicating that rating in ``Yes/No'' and free-form explanation generation are relatively easy tasks for the LLMs.
For citation, our REC models perform the best in both completeness and faithfulness metrics, while in instruction-following, they are ranked second to llama-3.1-70B and Claude3-Opus. 
Specifically, for the instruction-following metric, we find REC models often cite slightly more  sentences to ensure a good coverage of the entire conversation between a customer and an agent. In contrast, Llama-70B and Claude3 seem to be more selective.

\begin{table*}[t]
\small
\centering
\begin{tabular}{lccccc}
\hline
 & \textbf{Chat} & \textbf{Chat Hard} & \textbf{Safety} & \textbf{Reasoning} & \textbf{AVERAGE} \\
\hline
\textbf{Mistral-7B} & 76.10 & 54.30 & 81.50 & 53.90 & 66.50 \\
\textbf{Mistral-Nemo} & 86.60 & 59.60 & 74.80 & 70.30 & 72.80 \\
\textbf{Llama-3.1-70B} & 97.60 & 58.90 & 73.00 & 78.50 & 77.00 \\
\textbf{Claude-3-Opus} & 94.70 & 60.30 & 86.60 & 78.70 & 80.10 \\
\textbf{GPT-3.5} & 92.20 & 44.50 & 65.50 & 59.10 & 65.30 \\
\textbf{GPT-4} & 95.30 & 74.30 & 87.60 & 86.90 & 86.00 \\
\textbf{GPT-4o} & 96.10 & 76.10 & 88.10 & 86.60 & 86.70\\
\textbf{REC-8B} & 92.18 & 85.96 & 88.24 & 91.66 & 89.51\\
\textbf{REC-12B} & 91.90 & 86.60 & 92.00 & 93.60 & 91.00 \\
\textbf{REC-70B} & 94.10 & 90.10 & 93.20 & 96.40 & \textbf{93.50} \\
\hline
\end{tabular}
\caption{General LLM capabilities evaluation on RewardBench. Scores for all the compared LLMs are obtained from the RewardBench Leaderboard~\cite{RewardBench}.}
\label{tab:RewardBench}
\end{table*}

\begin{table*}[t]
\small
\centering
\begin{tabular}{lcccc}
\hline
\textbf{} & \textbf{LLM-FactVerify} & \textbf{Wiki-FactVerify} & \textbf{Summarization} & \textbf{Long-form QA} \\
\hline
\textbf{Mistral-7B} & 68.04 & 56.82 & 55.80 & 62.80 \\
\textbf{Mistral-Nemo} & 70.18 & 65.27 & 62.14 & 60.88 \\
\textbf{Llama-3.1-70B} & 75.25 & 68.84 & 69.56 & 76.98 \\
\textbf{Claude-3-Opus} & 76.82 & 76.52 & 70.65 & 76.82 \\
\textbf{GPT-3.5} & 72.81 & 69.53 & 68.60 & 70.40 \\
\textbf{GPT-4} & 77.44 & 77.22 & 72.45 & 76.92 \\
\textbf{GPT-4o} & 76.31 & 79.48 & 72.18 & 75.22 \\
\textbf{REC-8B} & 78.04 & 72.46 & 66.27 & 72.73 \\
\textbf{REC-12B} & \underline{78.78} & \underline{85.34} & \underline{72.51} & \underline{77.23} \\
\textbf{REC-70B} & \textbf{79.67} & \textbf{87.48} & \textbf{74.04} & \textbf{79.12} \\
\hline
\end{tabular}%
\caption{LLM-AggreFact performance across four common use-cases: LLM-FactVerify (ClaimVerify + FactCheck + Reveal), Wiki-FactVerify (WiCE), Summarization (AggreFact + TofuEval), and Long-form QA (ExpertQA + LFQA). The best and the second best performance for each metric are in bold and underlined separately.}
\label{tab:llm-aggrefact}
\end{table*}

\begin{table*}[t]
\small
\centering
\resizebox{\textwidth}{!}{%
\begin{tabular}{lcccccccc}
\hline
 & \textbf{Order} & \textbf{Bandwagon} & \textbf{Compassion} & \textbf{Selective} & \textbf{Salience} & \textbf{Distraction} & \textbf{Frequency} & \textbf{AVERAGE} \\
\hline
\textbf{Mistral-7B} & 0.163 & 0.518 & 0.346 & 0.671 & 0.603 & 0.0143 & 0.1406 & 0.3508 \\
\textbf{Mistral-Nemo} & 0.126 & 0.543 & 0.238 & 0.725 & 0.595 & 0.0196 & 0.0059 & 0.3218 \\
\textbf{Llama-3.1-70B} & 0.115 & 0.251 & 0.318 & 0.545 & 0.599 & 0.0116 & 0.0705 & 0.2729 \\
\textbf{Claude-3-Opus} & 0.078 & 0.081 & 0.380 & 0.483 & 0.589 & 0.0093 & 0.0087 & 0.2327 \\
\textbf{GPT-3.5} & 0.164 & 0.877 & 0.371 & 0.851 & 0.644 & 0.0029 & 0.0183 & 0.4183 \\
\textbf{GPT-4} & 0.092 & 0.098 & 0.399 & 0.431 & 0.569 & 0.0015 & 0.0046 & 0.2279 \\
\textbf{GPT-4o} & 0.075 & 0.086 & 0.417 & 0.468 & 0.578 & 0.0089 & 0.0112 & 0.2349 \\
\textbf{REC-8B} & 0.105 & 0.273 & 0.311 & 0.582 & 0.590 & 0.0119 & 0.0052 & 0.2683\\
\textbf{REC-12B} & 0.089 & 0.225 & 0.291 & 0.526 & 0.583 & 0.0108 & 0.0048 & 0.2471 \\
\textbf{REC-70B} & 0.073 & 0.095 & 0.258 & 0.487 & 0.572 & 0.0091 & 0.0043 & \textbf{0.2141} \\
\hline
\end{tabular}
}
\caption{Bias evaluation on CoBBLEr (lower is better).}
\label{tab:CoBBLEr}
\end{table*}

\subsection{General Capabilities}
\paragraph{RewardBench}
The RewardBench dataset assesses LLMs' performance across several critical abilities using curated chosen-rejected response pairs~\cite{RewardBench}. The evaluation involves determining the model's win percentage based on its ability to score the ``chosen'' response higher than the ``rejected'' one. Specifically, it's focused on testing chat performance, safety measures, and reasoning~(both in terms of code and math skills), while controlling for potential biases such as overfitting to prior datasets.

We first create prompts following the RewardBench evaluation format, where each prompt consisting of a query and two potential responses. The task for LLMs is to evaluate the two responses and determine which is better. The performance of the reward models is quantified by calculating the win percentage for chosen-rejected pairs associated with each prompt. A ``win'' is defined by a scenario where an LLM presents a preference for the selected answer. This quantitative measure provides a clear benchmark for evaluating model performance across tasks. By averaging the results across the 4 core sections: Chat, Chat Hard, Safety, and Reasoning, a comprehensive measure of each model's win percentage is derived.
As shown in Table~\ref{tab:RewardBench}, our REC models are able to achieve SOTA performance among all generative LLMs on the RewardBench Leaderboard, demonstrating their outstanding ability to serve as an auto-evaluator for pairwise comparisons.

\paragraph{LLM-AggreFact}
This is a benchmark for measuring the grounding capabilities of auto-evaluators. Given a reference document and a claim, the auto-evaluator determines if the claim is fully supported by the document. This holistic benchmark combines 10 attribution datasets used in recent studies on LLM factuality. 
Given some of the models such as GPT-3.5 have smaller context windows, we removed examples longer than 16K tokens from our evaluation. We use the same prompt instructions across categories, as shown in Appendix~\ref{appendix:LLM-Aggrefact-prompt}. To reduce model API costs, we randomly sampled 256 examples per evaluation task similar to the approach used in the FLAMe~\cite{vu2024foundational}. 

Table~\ref{tab:llm-aggrefact} presents our attribution results on LLM-AggreFact~\cite{tang2024minicheck}, categorized into four common use-cases: (1) LLM-FactVerify: fact verification of LLM-generated responses, (2) Wiki-FactVerify: evaluating correctness of Wikipedia claims, (3) Summarization: assessing faithfulness of summaries, and (4) Long-form QA: evaluating long-form answers to questions. REC-12B and REC-70B outperform all other models in all four categories, and REC-8B outperform all the other similar-sized models~(Mistral-7B and Mistral-Nemo), demonstrating their strong performance in attribution on diverse datasets.

\subsection{Bias Testing}
\paragraph{CoBBLEr}
Recent studies have found that LLM-as-a-Judge often exhibits cognitive biases, such as preferences for verbosity, egocentrism, bandwagon, and an overly authoritative tone~\cite{wang-etal-2024-large-language-models-fair,koo-etal-2024-benchmarking,zheng:23,Chen2024HumansOL}. To investigate the biases of the compared models, we evaluate them on the CoBBLEr benchmark~(Cognitive Bias Benchmark for LLMs as EvaluatoRs) \cite{koo-etal-2024-benchmarking}. This dataset is designed to evaluate the quality and reliability of LLMs when used as automated evaluators in a question-answering~(QA) setting. It assesses the presence of six cognitive biases, both implicit and induced, when LLMs are tasked with ranking responses generated by various other models. CoBBLEr's core objective is to identify the extent of bias in LLM evaluation outputs.

The CoBBLEr dataset includes 50 QA instructions, randomly selected from two well-established benchmarks: BIG-bench~\cite{srivastava2023beyond} and ELI5~\cite{fan-etal-2019-eli5}. 16 LLMs, both open and closed-source models, generate responses to these instructions. The evaluations involve pairwise comparisons between the responses of two models, wherein each model also acts as an evaluator to rank its own and others' outputs. The biases tested are categorized into two groups: (1) Implicit biases, such as egocentric bias~(where a model tends to prefer its own outputs), and (2) Induced biases, such as order bias, where the ranking of responses is influenced by their order in the evaluation.

Table~\ref{tab:CoBBLEr} presents the performance of the compared models on the CoBBLEr benchmark with metrics of order bias, bandwagon effect, compassion, selective bias, salience, distraction, and frequency. Note that lower scores indicates better performance ~(i.e., fewer biases) on CoBBLEr.
Overall, our REC-70B model has the best average performance with a score of 0.2141, followed closely by GPT-4~(0.2279) and GPT-4o~(0.2349). 
REC-12B and REC-8B, on the other hand, outperforms all the other same-sized and smaller models (Mistral-Nemo and Mistral-7B), and even some larger models (Llama-3-70B and GPT-3.5). The results suggest that training with REC produces an LLM judge with fairer evaluation capabilities, independent of the cognitive biases tested, such as verbosity, order, or an overly authoritative tone. In contrast, off-the-shelf LLMs can be more influenced by factors such as the order of responses or the length of the text.

\section{Conclusions}
LLMs excel in generating coherent and high-quality text for various tasks, but evaluating content quality is challenging due to issues like factual inaccuracies and hallucination. This paper introduces three fine-tuned models—REC-8B, REC-12B, and REC-70B—designed to evaluate generated text across several dimensions: faithfulness, instruction following, coherence, completeness. These models not only provide ratings for these metrics but also offer detailed explanations and verifiable citation, i.e., Ratings, Explanations, and citation (REC), thereby enhancing trust in the content. To achieve this goal, a curated dataset for explanations and citation fine-tuning is proposed to facilitate future research, which is the first public dataset containing both content quality citation and RAG citation. The models support various citation modes, including Post-fix citation, Post-fix citation mode with snippet, Inline citation, and Inline citation mode with snippet to balance the trade-off between latency and granularity of citation. Extensive evaluations on diverse benchmarks (i.e., ALCE, ExpertQA, ABCD summarization, RewardBench, LLM-AggreFact, and CoBBLEr) demonstrate that the fine-tuned LLM auto-evaluator, REC-70B, outperforms state-of-the-art models. It excels in content evaluation, delivering clear explanations and ensuring citation accuracy.

\section*{Acknowledgements}
We appreciate Yilun Zhou, Shafiq Rayhan Joty, Peifeng Wang, and Austin Xu from Salesforce AI research for insightful discussions and help to benchmark model performance. We gratefully acknowledge partial support from NSF grant DMS-2413265, NSF grant 2023505 on Collaborative Research: Foundations of Data Science Institute (FODSI), the NSF and the Simons Foundation for the Collaboration on the Theoretical Foundations of Deep Learning through awards DMS-2031883 and 814639, NSF grant MC2378 to the Institute for Artificial CyberThreat Intelligence and OperatioN (ACTION), and a Berkeley Deep Drive (BDD) grant from BAIR and a Dean's fund from CoE, both at UC Berkeley.

\section*{Limitations}
Our work introduces the innovative \textbf{R}ate, \textbf{E}xplain, and \textbf{C}ite~(\textbf{REC}) model for trusted automated AI evaluations, offering significant benefits but with a few limitations that future research will address. First, our work is limited to the baselines we compare, as non-LLM citation methods, for instance, approaches that leverage sentence embedding \cite{reimers2019sentencebertsentenceembeddingsusing} similarity to identify and cite the most semantically relevant sources were not included. Including such techniques could serve as valuable baselines to further validate our model's generalization and effectiveness. Second, the need for post-generation explanations and citation may introduce latency. Future work will compare latencies across models and explore optimizations to improve speed. We are also considering fine-tuning the snippet output using starting and ending character numbers to optimize for latency.  Finally, even though the 12B base model is capable of generating multi-lingual output, our current evaluations are limited to English datasets. Expanding to multilingual evaluations is a key goal for future research.

\section*{Impact Statement}
Our contributions present a novel general-purpose auto-evaluator that supports various modes of citation, as well as a dataset with citation and explanations to support further research. There are a few ethical considerations for users of such technology.
As with the use of fine-tuning and RAG, it is important to be aware of potential propagating biases and feedback loops. Biased data used for training and/or retrieval could lead to model outputs that reinforces such biases, impacting model quality and fairness. 
Although we extensively evaluated our LLM auto-evaluator and found it to be safe, unbiased, and factual, users should consider whether such considerations hold true for the input source itself. Of note, even established sources such as knowledge banks may contain inaccurate or inconsistent information. 
Finally, though we aimed to be transparent and consistent for our data labeling tasks, using clear instructions and validations, note that data labeling tasks are inherently subject to the bias of labeler background and experiences.


\bibliography{example_paper}
\bibliographystyle{icml2025}

\newpage
\appendix

\section{Content Quality Citation Modes and the Evaluation Prompt}
\label{sec:appendix_content_quality_citation}
Here we provide the evaluation prompt template used in Figure~\ref{fig:abcd_illustration_rec}, the raw auto-evaluator output, and a final output example in an alternative citation mode for content quality evaluation.

\subsection{Content Quality Evaluation Prompt}
\begin{small}\begin{verbatim}
You will be asked to provide 
feedback on one quality metric for a task
prompt given to another LLM and that 
LLM's output. Given the task prompt as a 
context, your job is to give feedback on 
whether there are any errors in the output,
based on the quality metric. If there 
are no errors in the output, please 
also provide your feedback that there are
no errors. Remember your main goal is to
provide feedback so that another LLM can
use your feedback to improve its
generation.
You need to support statements in your 
feedback that can be linked back to ANOTHER 
LLM'S TASK PROMPT with citations.
Please make sure you read and understand 
these instructions carefully. Please keep 
this document open while reviewing, and 
refer to it as needed.
Please only generate your answer following
the Output JSON format. Do not generate 
anything else.

Evaluation criteria:
{metric_name} ({metric_scale})
- {metric_description}

Respond using the following JSON format
based on the provided ANOTHER LLM'S
TASK PROMPT and ANOTHER LLM'S OUTPUT:
Desired format:
{
   "answer":<answer>,
   "feedback":<generated feedback>,
   "statements":[
   {
       "statement_string":<one statement 
       extracted from feedback>,
       "citations":[
       {
            "snippet":<snippet>
       }]
    }]
}

### INSTRUCTIONS:
Here are the instructions to follow to 
generate the response.

1. Generate a feedback based on ANOTHER 
LLM'S TASK PROMPT and ANOTHER LLM'S 
OUTPUT. Explain whether the evaluation 
criteria is met.
    - Set "answer" to "Yes" if the
    evaluation criteria is met, 
    and "No" otherwise.
    - Set "feedback" to the 
    generated feedback.
2. For each statement in your feedback, 
extract relevant sentences that support
the statement from ANOTHER LLM'S
TASK PROMPT.
3. For each extracted sentence:
    - Set "snippet" to be the extracted 
    sentence.
4. Compile relevant snippets for the 
statement:
    - Set "statement_string" to be 
    a statement in the generated feedback.
    - Set "citations" to be a list of JSON 
    objects containing all relevant
    snippets (from step 3) supporting the 
    "statement_string".
5. Compile the statements into a list of 
JSON objects.
    - Set "statements" to be a list of 
    JSON objects containing metadata of 
    all statements (from step 4) in the 
    feedback.

[BEGIN DATA]
### ANOTHER LLM'S TASK PROMPT:
Summarize knowledge from transcripts
after they've ended. Summarizing the key
points of the conversation including
customer issue and resolution.

### Conversation:
{convo}

### ANOTHER LLM'S OUTPUT:
{answer}
[END DATA]

### Response(JSON only): 
\end{verbatim}\end{small}

\subsection{Content Quality Citation Modes}
Using the same input example as in Figure~\ref{fig:abcd_illustration_rec}, we show the raw auto-evaluator output that allows users to further process according to either of the two citation modes, ``post-fix with context snippet'' and ``inline with context snippet'', to fit their needs. With the final output in ``inline with context snippet'' mode already presented in Figure~\ref{fig:abcd_illustration_rec}, here we show another possible final output format in ``post-fix with context snippet'' mode.

\subsubsection{Raw auto-evaluator Output}
Citations are extracted verbatim from the input context to support statements in the auto-evaluator's explanation.

\begin{small}\begin{verbatim}
"statements": [
 {
   "statement_string": "The answer covers
   all the key points of the conversation,
   including the customer's question, the
   agent's response, and the resolution.",
   "citations": ["Summarizing key points
   of the conversation including customer
   issue and resolution.", "Customer: Is it
   possible to have my items gift wrapped?"
   ,"Agent: We offer gift wrapping but it
   cost $4.99 per item"]
 },
 {
   "statement_string": "It also mentions the
   customer's reason for asking about gift
   wrapping and the agent's polite closing."
   ,"citations": ["Customer: Oh great. My
   Bestie's birthday is coming up, that's
   the reason for my questions. Thanks",
   "Agent: All right. thanks for
   reaching out and have a great day!"]
  }
]
\end{verbatim}\end{small}

\subsubsection{Final Output Example in ``post-fix with context snippet'' Citation Mode}
\begin{small}\begin{verbatim}
The answer covers all the key points of the
conversation, including the customer’s
question, the agent’s response, and the
resolution. It also mentions the customer’s
reason for asking about gift wrapping and
the agent’s polite closing.
[1][2][3][4][5]

[1]: "Summarize key points of the
conversation including customer issue
and resolution."
[2]: "Customer: Is it possible to have
my items gift wrapped?"
[3]: "Agent: We offer gift wrapping but
it costs $4.99 per item." 
[4]: "Customer: Oh great. My Bestie’s
birthday is coming up, that’s the reason
for my questions. Thanks."
[5]: "Agent: All right. thanks for
reaching out and have a great day!"

\end{verbatim}\end{small}

\section{Details of using REC Models for General RAG citations}
\label{sec:appendix_rag_citation}

\begin{figure*}[t]
    \centering
    \includegraphics[width=\textwidth, height=\textheight, keepaspectratio]{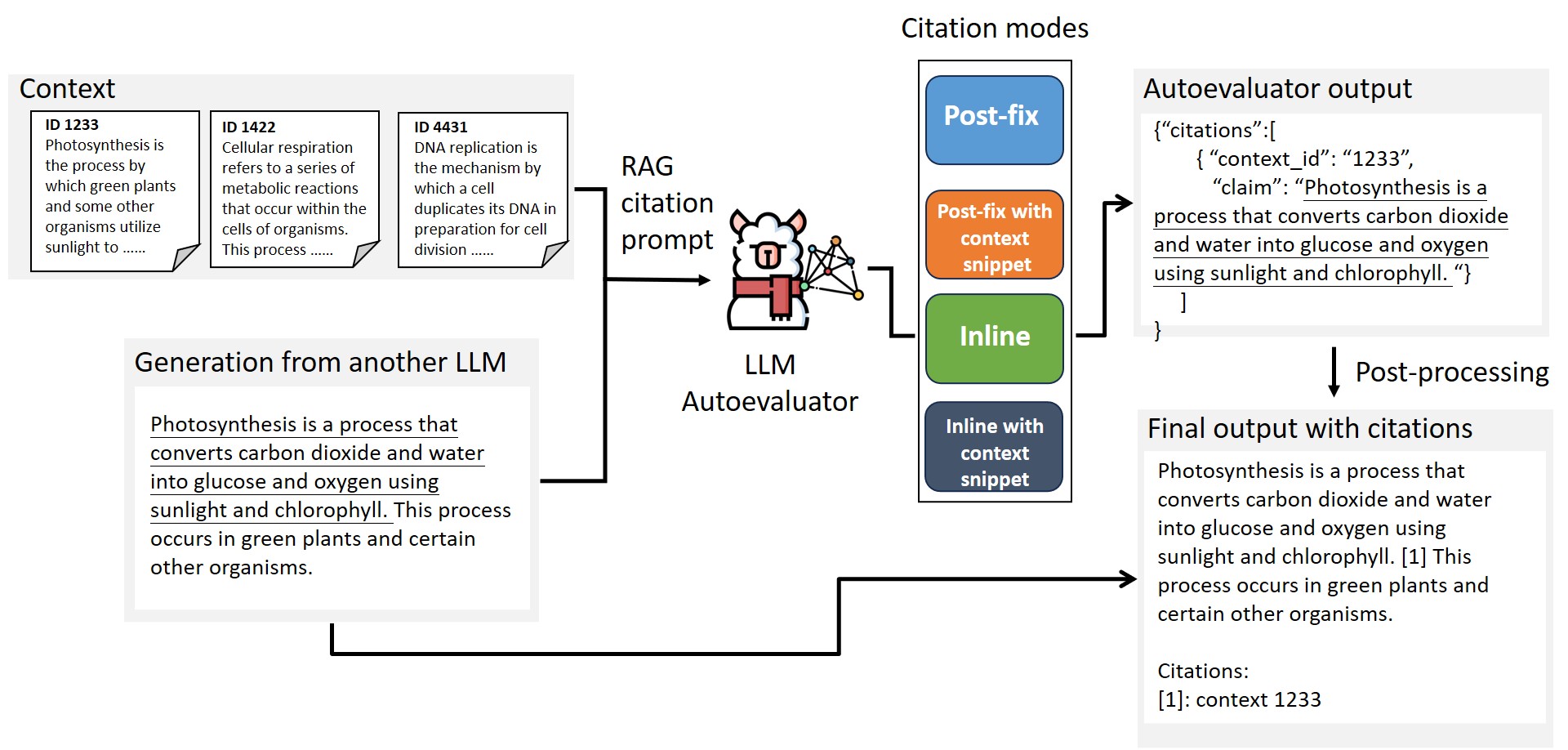}
    \caption{Illustration of using REC models for general RAG citations.}
    \label{fig:rag_illustration}
\end{figure*}

We demonstrate how REC models can also be used for general RAG citations in Figure~\ref{fig:rag_illustration}. In the example shown, we present this question ``What is photosynthesis?'' to another LLM, and collect retrieved articles with ids as the context, and the generated answer from another LLM to serve as inputs. The goal is for our LLM auto-evaluator to provide citations extracted verbatim from the context to support the generation from another LLM. REC models support four different citation modes, and the details can also be found in this section.

\begin{itemize}
\item \textbf{Retrieved Context}:

    ID 1233
    
    Photosynthesis is the process by which green plants and some other organisms utilize sunlight to synthesize their food. This remarkable process involves the conversion of carbon dioxide and water into glucose and oxygen, facilitated by chlorophyll. It serves as the foundation for energy production in these organisms and plays a crucial role in maintaining the balance of oxygen in our atmosphere.

    ID 1422
    
    Cellular respiration refers to a series of metabolic reactions that occur within the cells of organisms. This process transforms biochemical energy from nutrients into adenosine triphosphate (ATP), which is essential for cellular functions. During cellular respiration, waste products are released, and oxygen is a vital element needed for the process to function efficiently. This energy conversion is critical for sustaining life.

    ID 4431
    
    DNA replication is the mechanism by which a cell duplicates its DNA in preparation for cell division. This process ensures that genetic information is accurately copied and passed on to daughter cells. The enzyme DNA polymerase is instrumental in DNA replication, playing a crucial role in maintaining genetic fidelity and continuity across generations.

\item \textbf{LLM Generated Response to Cite}:

Photosynthesis is a process that converts carbon dioxide and water into glucose and oxygen using sunlight and chlorophyll. This process occurs in green plants and certain other organisms.
\end{itemize}

\subsection{RAG Citation Prompt}
As an example, below is the prompt that we used to generate inline citations. This can be modified to generate responses in each of the different RAG citation modes shown in the following sections.

\begin{small}\begin{verbatim}
Your task is to provide
citations for a generated response
from an LLM for a RAG 
(Retrieval Augmented Generation)
application. You will be provided two 
sections below. The RETRIEVED CHUNKS
section is the context from 
where you will generate citations.
The LLM GENERATED ANSWER section
is for the LLM generated answer
from where you will generate claims.

###
Respond using the following JSON format:

Desired format:

{
   "citations":[
      {
         "context_id":"<context_id>",
         "claim": "<claims identified from 
         LLM GENERATED ANSWER>"
      }
   ]
}

###INSTRUCTIONS
In order to generate citations, 
follow these steps below:

1. Review the "LLM GENERATED ANSWER" 
section below and extract a set of claims,
VERBATIM and set the "claim" in the output
json. 
2. Your task is to generate  a citation
for each of the claims identified 
in Step 1, by following the steps below:
    i. For each claim, identify the
    most relevant chunk and the
    corresponding "context_id" from
    the "RETRIEVED CHUNKS" section that
    supports the claim. 
    ii. For the identified chunk, 
    set the "context_id" in the output
    to the corresponding context_id. 
    Set "context_id" to "None" if none
    of the chunks in the
    "RETRIEVED CHUNKS" section supports
    the claim. 
3. Ensure that you structure your 
response to adhere to the desired
output JSON format.

###
RETRIEVED CHUNKS:
{retrieved_chunks}

###
LLM GENERATED ANSWER:
{answer}

###
Response (JSON only):
\end{verbatim}\end{small}

\subsection{Post-fix Citation}
\begin{small}\begin{verbatim}
{
  "citations": [
    {
      "context_id": "1233"
    }
  ]
}
\end{verbatim}\end{small}

\subsection{Post-fix Citation with Context Snippet}
\begin{small}\begin{verbatim}
{
  "citations": [
    {
      "context_id": "1233",
         "snippet": "Photosynthesis is
         the process by which green
         plants and some other organisms
         utilize sunlight to synthesize
         their food. This remarkable
         process involves the 
         conversion of carbon
         dioxide and water into
         glucose and oxygen, 
         facilitated by chlorophyll"
    }
  ]
}
\end{verbatim}\end{small}

\subsection{Inline Citation}
\begin{small}\begin{verbatim}
{
  "citations": [
    {
      "context_id": "1233",
         "claim": "Photosynthesis is a
         process that converts carbon
         dioxide and water into glucose
         and oxygen using sunlight and 
         chlorophyll."
    }
  ]
}
\end{verbatim}\end{small}

\subsection{Inline Citation with Context Snippet}
\begin{small}\begin{verbatim}
{
  "citations": [
    {
      "context_id": "1233",
       "claim": "Photosynthesis is a
         process that converts carbon
         dioxide and water into glucose
         and oxygen using sunlight and 
         chlorophyll.",
       "snippet": "Photosynthesis is the 
         process by which green 
         plants and some other 
         organisms utilize sunlight 
         to synthesize their food. This
         remarkable process involves
         the conversion of carbon
         dioxide and water into
         glucose and oxygen, 
         facilitated by chlorophyll"
    }
  ]
}
\end{verbatim}\end{small}

\section{REC-Data Details}
\label{appendix:rec-data}

For general instruction tasks, we include summarization on ABCD~\citep{chen-etal-2021-action}, a dataset containing customer support dialogue transcripts, in five languages.
For pointwise evaluation, we leverage HelpSteer2~\citep{helpsteer2} and perform synthetic data generation~(Section~\ref{sec:synthetic-data-gen}) on the (dialogue, summary) pairs from ABCD, and QA pairs from FloDial.
For pairwise and open-ended evaluation tasks, we use Auto-j~\citep{li2023generative-auto-j}, Shepherd~\citep{wang2023shepherd}, Skywork~\footnote{\url{https://huggingface.co/datasets/Skywork/Skywork-Reward-Preference-80K-v0.1}}, HelpSteer2\footnote{\url{https://huggingface.co/datasets/nvidia/HelpSteer2}}, OffsetBias\footnote{\url{https://huggingface.co/datasets/NCSOFT/offsetbias}}, and Code Preference\footnote{\url{https://huggingface.co/datasets/Vezora/Code-Preference-Pairs}}.
For citations, we perform synthetic data generation~(Section~\ref{sec:synthetic-data-gen}) on FloDial for both RAG citations and content quality citations, and on ABCD, Auto-j, Shepherd, and Email Thread Summary~\footnote{\url{https://www.kaggle.com/datasets/marawanxmamdouh/email-thread-summary-dataset/data}} for content quality citations.

\begin{figure*}[ht]
    \centering
    \includegraphics[width=0.8\textwidth, height=0.8\textheight, keepaspectratio]{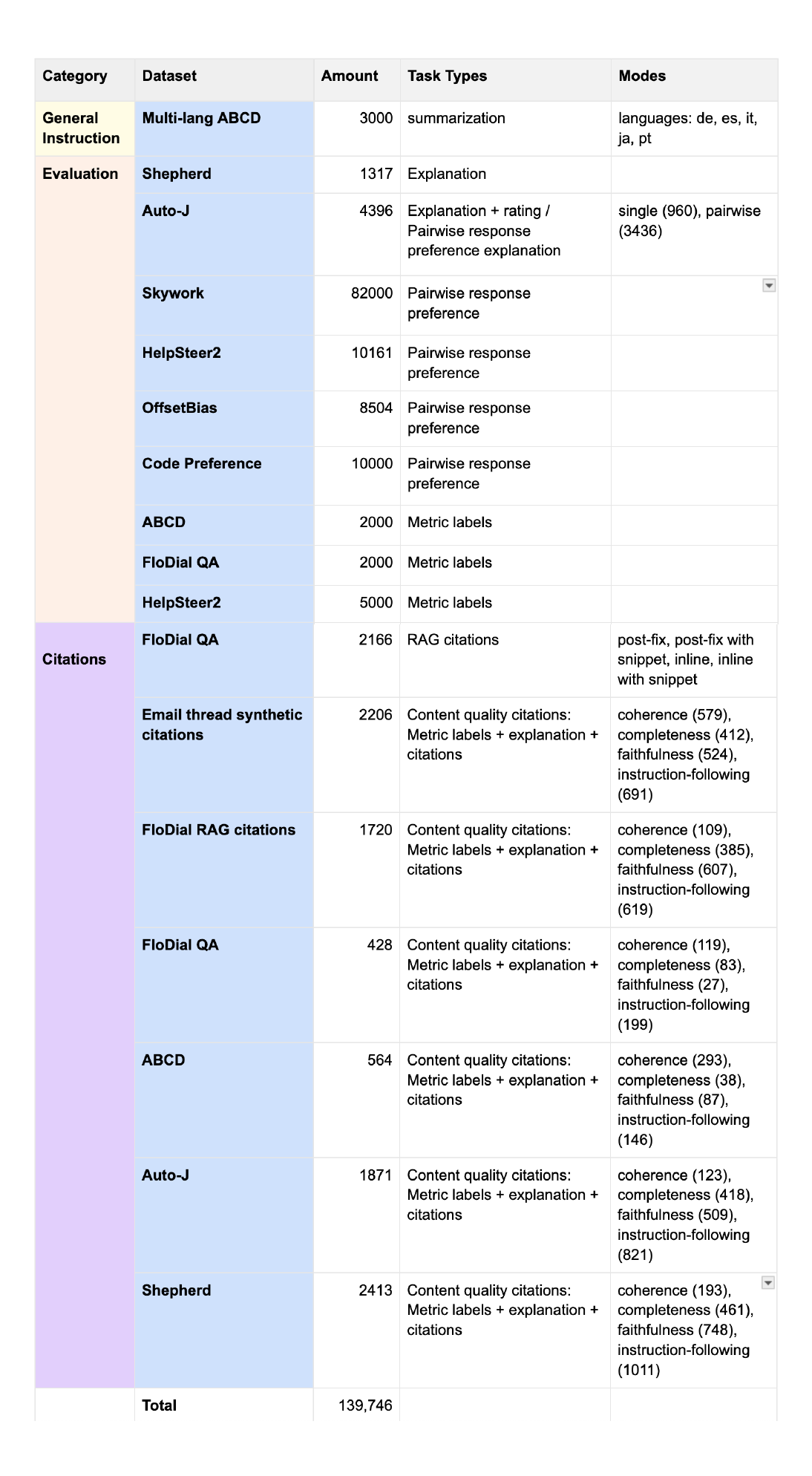}
    \caption{Detailed breakdown of REC-Data distribution.}
    \label{fig:rec-data}
\end{figure*}

Note that it's possible to derive different tasks from the same dataset when different task prompts are assigned to an LLM. We provide a detailed breakdown of REC-Data in Figure~\ref{fig:rec-data}.

\section{Synthetic Data Generation Prompt Example}
\label{appendix:synthetic_prompt}
The prompts used to generate synthetic content quality and RAG citation data are the same as the ones shown in Appendices~\ref{sec:appendix_content_quality_citation} and~\ref{sec:appendix_rag_citation}.
\subsection{Pointwise Evaluation ("Faithfulness metric")}
\begin{small}\begin{verbatim}
You will be given a source text given to
another LLM, and that LLM's answer.
Your task is to rate that LLM's answer 
on one quality metric.
Please make sure you read and understand 
these instructions carefully. Please 
keep this document open while reviewing, 
and refer to it as needed.


Evaluation criteria:
metric_name = 'Faithfulness'
metric_scale = 'Yes/No'
metric_description = ""
The generated answer only contains truthful
content, and does not contain invented 
or misleading facts that are 
not supported by the context.
""


Evaluation steps:
1. Read the source text.
2. Read the other LLM's answer and compare 
it to the source text. Check if the 
evaluation criteria is met.
3. If the evaluation criteria is met, 
answer Yes; if the evaluation criteria 
is not met, answer No.
4. Create a JSON response
{
    "metriclabel": <Yes|No only>,
    "justification": 
    <explanation for the score>
}

Example:
Source text given to another LLM:
{query_with_context}

The other LLM's answer:
{answer}

### Response(JSON Only):
\end{verbatim}\end{small}

\section{Remark on Why The Three Metrics in ALCE are All Important}
\label{sec:appendix_alce_remark}
Regarding why the three criteria are \emph{equally important}, we encourage the reader to consider from an \emph{user’s perspective}. An LLM’s citation response cannot be deemed good if it is lacking in any of the three criteria: fluency, answer correctness, and citation accuracy. To illustrate this, we aim to provide an example of a good citation response that scores well on all three criteria, as well as three examples that each fail in one of the criteria (assuming the same RAG QA setting as in Appendix~\ref{sec:appendix_rag_citation}).

\begin{itemize}
    \item \textbf{Example 1 (scoring high in all criteria):} Photosynthesis is the process by which green plants and some other organisms utilize sunlight to synthesize their food. This remarkable process involves the conversion of carbon dioxide and water into glucose and oxygen, facilitated by chlorophyll [1233].

    \item \textbf{Example 2 (lacks fluency):} Photosynthesis, it's the way that green plants and also some other organisms they use (*not fluent) sunlight for making their food. This process, it's interesting, converts carbon dioxide with water into glucose and oxygen, and chlorophyll helps with it [1233].

    \item \textbf{Example 3 (lacks correctness):} Photosynthesis, it's when green plants and other organisms use water (*wrong) to directly create sunlight (*wrong) as food. This process, it's interesting, turns oxygen into glucose and carbon dioxide, with chlorophyll doing nothing important.

    \item \textbf{Example 4 (lacks citation accuracy):} Photosynthesis is the process by which green plants and some other organisms utilize sunlight to synthesize their food. This remarkable process involves the conversion of carbon dioxide and water into glucose and oxygen, facilitated by chlorophyll [1422] (*wrong citation).
\end{itemize}
It is highly likely that a user would prefer the first example, which is well-balanced across all three criteria, as a good response. We hope this illustration better conveys our point.

\section{ABCD Content Quality Citation Human Evaluation Instruction}
\label{sec:appendix_human_label}
Annotation instruction:
You will be evaluating attributed judgments made by an LLM auto-evaluator for answers generated by another LLM against three metrics: instruction-following, completeness, and faithfulness.
Your task is to evaluate LLM auto-evaluator’s answer by completing the following THREE tasks: *you can skip if the output shows ``invalid json''.
\begin{itemize}
    \item \textbf{Metric answer correctness:} Please label ``yes''/``no'' by evaluating whether the metric ``answer'' in auto-evaluator’s output is correct or not, based on the provided conversation, task prompt, and summary.

    \item \textbf{Metric explanation correctness:} Please label ``yes''/``no' by evaluating whether the ``feedback'' in auto-evaluator’s output is correct or not, based on the provided conversation, task prompt, and summary. Only proceed to citation annotation (next task) if you label ``yes'' in this task, namely the explanation is correct.

    \item \textbf{Citation annotation:} In ``statements'', you will find different ``statement\_string'' extracted from the ``feedback''. Please provide citation for each ``statement\_string'' by directly copying relevant sentences (Please cite full sentences, not substrings) from ``task prompt'' and ``chat'' columns (Please DON’T cite from ``summary''). If the ``statement\_string'' is hallucinated and doesn’t make sense in the first place, please put ``[<halu>]'' as the citation to signal.
\end{itemize}

\section{Prompt for LLM-AggreFact Eval}
\label{appendix:LLM-Aggrefact-prompt}
\begin{small}\begin{verbatim}
Determine whether the provided claim is
consistent with the corresponding document.
Consistency in this context implies that
all information presented in the claim is
substantiated by the document. If not,
it should be considered inconsistent.

Document: {doc}
Claim: {claim}

Please assess the claim's consistency
with the document by responding with
either "yes" or "no".

Answer (yes|no only):
\end{verbatim}\end{small}



\end{document}